\documentclass{article}
\usepackage{spconf,amsmath,graphicx}
\usepackage{booktabs}
\usepackage{multirow}
\usepackage[vlined,linesnumbered,ruled]{algorithm2e}
\usepackage{xcolor}
\usepackage{amsfonts}
\usepackage{etoolbox}
\usepackage{enumitem}
\usepackage{caption}
\usepackage{url}
\usepackage{xcolor}
\usepackage{tikz}
\usepackage{hyperref}
\usepackage{mathtools}
\usepackage{graphicx}
\usepackage{float}

\newcommand*\circled[1]{\tikz[baseline=(char.base)]{
    \node[shape=circle,draw,inner sep=1pt] (char) {#1};}}
\setlist[itemize]{topsep=0pt, partopsep=0pt, parsep=0pt, itemsep=0pt}

\usepackage[
backend=biber,
style=ieee,
doi=false,isbn=false,url=false,eprint=false
]{biblatex}

\defbibheading{bibliography}[\refname]{}

\apptocmd{\thebibliography}{\setlength{\itemsep}{0.1pt}}{}{}

\title{FedMM: Federated Multi-Modal Learning with Modality Heterogeneity in Computational Pathology}

\name{Yuanzhe Peng, Jieming Bian, Jie Xu
\thanks{This work is supported in part by NSF under grants 2033681, 2006630, 2044991, 2319780. 
Email: \{ypeng, jxb1974, jiexu\}@miami.edu}}

\address{University of Miami, FL, USA}
\begin{document}
\ninept
\maketitle
\begin{abstract}
The fusion of complementary multimodal information is crucial in computational pathology for accurate diagnostics. However, existing multimodal learning approaches necessitate access to users' raw data, posing substantial privacy risks. While Federated Learning (FL) serves as a privacy-preserving alternative, it falls short in addressing the challenges posed by heterogeneous (yet possibly overlapped) modalities data across various hospitals.
To bridge this gap, we propose a Federated Multi-Modal (FedMM) learning framework that federatedly trains multiple single-modal feature extractors to enhance subsequent classification performance instead of existing FL that aims to train a unified multimodal fusion model. Any participating hospital, even with small-scale datasets or limited devices, can leverage these federated trained extractors to perform local downstream tasks (e.g., classification) while ensuring data privacy.
Through comprehensive evaluations of two publicly available datasets, we demonstrate that FedMM notably outperforms two baselines in accuracy and AUC metrics.

\end{abstract}

\section{Introduction}
\label{sec:intro}
Recent advancements in computational pathology have triggered a revolution in medical diagnostics and research \cite{vanguri2022multimodal}, \cite{lipkova2022artificial}. A critical aspect of this progress involves leveraging complementary data from diverse sources to improve diagnosis through machine learning \cite{luo2019enhancing}, \cite{anagnostou2020multimodal}. For instance, researchers can combine histology data, such as Whole Slide Images (WSI), with genomic data like Copy Number Variations (CNV) to gain a comprehensive understanding of cancer subtypes \cite{wang2021lung}. This multimodal approach leads to enhanced diagnostic accuracy and personalized treatment options \cite{gao2020mgnn}.

However, multimodal data fusion raises significant privacy concerns, particularly when handling medical data \cite{nguyen2022federated}. To tackle this issue, Federated Learning (FL) \cite{mcmahan2017communication}, a privacy-conscious distributed learning framework, has drawn substantial attention. FL facilitates the local update of models, which are then uploaded to a central server, thus eliminating the requirement to share raw data \cite{adnan2022federated}. Although current FL methods prove effective in scenarios with single-modal FL or multimodal FL with homogeneous modality distributions (i.e., the input maintains a consistent structure), 
they fall short in multimodal FL systems with modality heterogeneity \cite{qu2022rethinking}.
\begin{figure}
\centering
  \includegraphics[width=0.54\textwidth, trim=170 130 100 80,clip]{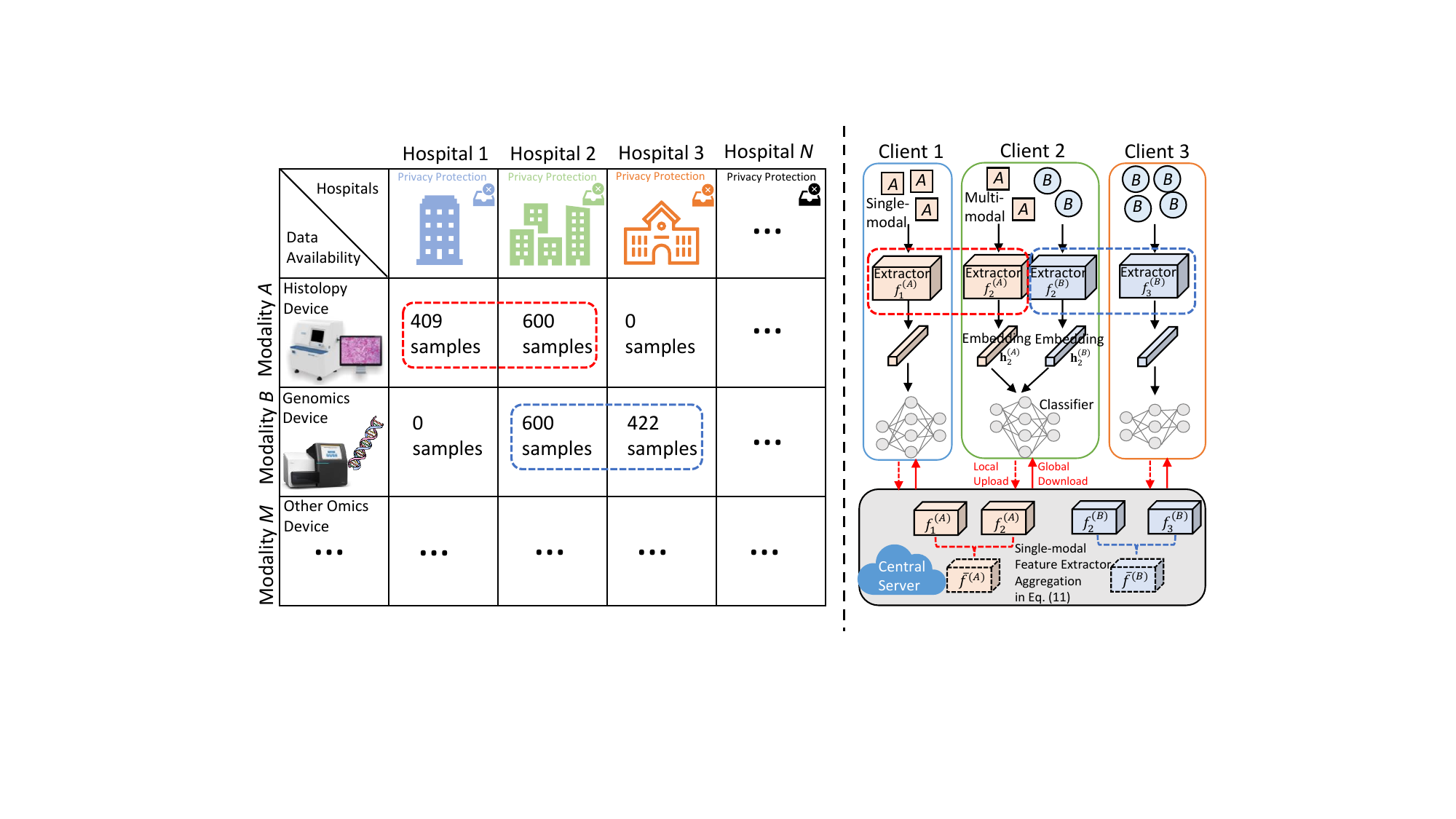}
  \caption{The problem of modality heterogeneity in computational pathology (left) and our solution (right). 
  The aim is to harness distributed privacy-sensitive data, which might overlap on some modalities across several hospitals, to train multiple single-modal feature extractors federatedly. These federated extractors are intended to exhibit superior performance compared to locally trained extractors.
  }
  \vspace{-4mm}
  \label{fig:1_Intro}
\end{figure}

This modality heterogeneity arises from multiple factors. 
First, equipment limitations at different hospitals may restrict the use of multiple diagnostic devices, leading to a narrow range of available data modalities for diagnosis, as illustrated in Fig.~\ref{fig:1_Intro}. Second, patient preferences for specific hospitals may result in some hospitals possessing limited available data modalities for diagnosing the same disease. Lastly, data housed in various hospitals may be non-identically and independently distributed (non-i.i.d). 
Consequently, collaborative training becomes unfeasible under the conventional FL framework. Hospitals are thus constrained to local model training, overlooking the potential gains that could be achieved by leveraging overlapped modalities across various hospitals (i.e., clients in FL).

To bridge this gap, we propose a Federated Multi-Modal (FedMM) Learning framework. Unlike traditional FL, which aims to train a multimodal fusion model, FedMM trains multiple single-modal feature extractors federatedly.
This allows each participating hospital to locally leverage these federated feature extractors for feature extraction and subsequent classification tasks.
Besides, as these federated feature extractors are trained using distributed data, their capacity for generalization is significantly enhanced. Specifically, any hospital can benefit from this framework, no matter its device availability and sample size. For example, even newly-established hospitals, constrained by limited samples and devices, can leverage these federated extractors to aid local model training while ensuring data privacy.
Our key contributions are summarized as follows:
\begin{itemize}
\item We leverage multimodal data to enhance performance in classifying diverse subtypes of lung and kidney cancers.
\item We propose FedMM, armed with an innovative dynamic loss function, to address the issue of modality heterogeneity in computational pathology while ensuring data privacy.
\item Through comprehensive evaluation of two publicly available datasets, we demonstrate that FedMM significantly outperforms two established baseline methods.
\end{itemize}


\vspace{-2mm}
\section{Related work}
\label{sec:related}
In computational pathology, multimodal information fusion significantly improved diagnostic accuracy \cite{qian2021prospective}, \cite{peneder2021multimodal}. Wang et al. developed a hybrid network that combined image and genomics data to improve lung cancer subtype diagnosis \cite{wang2021lung}. Chen et al. applied multimodal deep learning to analyze pathology images and molecular profiles across 14 cancer types \cite{chen2022pan}. 
However, these methods commonly overlooked the privacy-sensitive nature of medical data. The difficulty in collecting large-scale data for model training further restricted their real-world applicability \cite{acosta2022multimodal}.

The advent of FL presents a promising distributed training framework that addresses privacy concerns across different institutions \cite{kalra2023decentralized}, \cite{ogier2023federated}. Lu et al. showed FL's effectiveness in computational pathology \cite{lu2022federated}, while Li et al. proposed a privacy-preserving framework for multi-site fMRI analysis \cite{li2020multi}. However, existing solutions focused on single-modal FL, struggling with heterogeneous data often found across different clients \cite{li2021fedbn}, \cite{li2022federated}, \cite{vahidian2023rethinking}. 
Very recently, Yu et al. introduced CreamFL \cite{yu2023multimodal}, which employs the contrastive representation ensemble to train a global model with the support of multiple clients, diverging from our emphasis on client-side performance. Chen et al. presented FedMSplit \cite{chen2022fedmsplit}, which modularized client models into shareable blocks to accommodate diverse sensor setups.
Concurrently to our work, Ouyang et al. proposed Harmony~\cite{ouyang2023harmony}, but in the fusion phase, 
limited samples impeded the ideal aggregation of multimodal client classifiers within the same cluster, owing to the necessity for consistent modality combinations.

To the best of our knowledge, we are the first to tackle the FL issue with modality heterogeneity in computational pathology.

\vspace{-2mm}
\section{Method}
\label{sec:method}
\subsection{Problem Formulation}
In a typical FL setup, multiple clients work with one central server to train a shared model, denoted as $\mathcal{F}(\cdot) = g\left[f(\cdot)\right]$. Here, $f(\cdot)$ is the feature extractor like a Convolutional Neural Network (CNN), and $g(\cdot)$ is the classifier like a Multilayer Perceptron (MLP). The goal is to train a global model $\mathcal{F}(\cdot)$ using distributed data from different clients and minimize the loss function, which can be formulated as:
\begin{equation}
\scriptsize
\begin{aligned}
\underset{\mathbf{w}_i}{\arg\min} \left\{ \frac{1}{N}\sum_{i=1}^{N} \mathcal{L}\left[\mathcal{F}_i(\mathbf{w}_i;\mathbf{X}_i), y_i\right] = \frac{1}{N}\sum_{i=1}^{N} \mathcal{L}\left[\mathbf{w}_i; g_i\left[f_i(\mathbf{X}_i)\right], y_i\right] \right\}
\end{aligned}
\label{eq:eq1}
\end{equation}
where $N$ is the total number of clients, $\mathbf{w}_i$ represents the learnable weights associated with the $i$-th client model, $\mathbf{X}_i$ is the input, $y_i$ is the corresponding label, and $\mathcal{L}$ denotes the loss function. This method works well in
single-modal FL or multimodal FL with homogeneous modality distributions (i.e., consistent input structure). 

However, it falls short when dealing with heterogeneous modality distributions across various clients because clients may possess varying numbers and types of modalities data. Such heterogeneity hampers the optimization of feature extractor $f(\cdot)$ and classifier $g(\cdot)$ using existing FL methods like FedAvg \cite{mcmahan2017communication}, FedProx \cite{li2020federated}, etc.

Thus, the intuition behind our study is to investigate the feasibility of aggregating feature extractors corresponding to the same modality across all clients. 
Our strategy aims to train multiple single-modal feature extractors federatedly by utilizing privacy-sensitive, small-scale data from various local clients.
Subsequently, each client deploys these federated extractors to extract features from the corresponding modality data locally. These extracted features then serve as the input for each client's classifier. 
Notably, the aggregation process excludes the participation of classifiers from different clients since the nature of heterogeneous inputs is unavoidable.

Specifically, we consider a multimodal FL system comprising $N$ clients and one central server. Each client possesses up to $M$ different modalities in their local data, where $M \geq 2$. For any arbitrary client $C_i$, where $1 \leq i \leq N$, the local sensitive-privacy dataset is denoted as $\left\{\mathbf{X}_i, y_i\right\}$. 
Here, $\mathbf{X}_i = \{\mathbf{x}_i^{(k)}\}_{k=1}^{M_i}$ contains the number of $M_i$ modalities. 
To put it more clearly, for the client $i$, the available modalities are denoted by \(\forall k \in [M_i]\), while for the server, they are \(\forall k \in [M]\), and $[M_i] \subseteq [M]$.

For the client $C_i$, the data of each modality is individually fed into the corresponding single-modal feature extractor ${f_i}^{(k)}(\cdot)$ to generate $M_i$ sets of feature embeddings:
\begin{equation}
\{\mathbf{h}_i^{(k)}\}_{k=1}^{M_i} = \{{f_i}^{(k)}[\mathbf{x}_i^{(k)}]\}_{k=1}^{M_i}
\end{equation}
Here, $\mathbf{h}_i^{(k)}$ represents the feature embedding 
extracted by ${f_i}^{(k)}(\cdot)$. 
Subsequently, these diverse extracted single-modal feature embeddings can be fused (concatenated) to combine complementary information, provided they originate from the same sample. The resulting concatenated multimodal feature embedding 
is then used as input to the local classifier \( g_i(\cdot) \) for making prediction $\hat{y}_i$, as expressed below (the concatenated expression $\oplus$ is omitted for brevity):
\begin{equation}
\hat{y}_i = g_i\{{f_i}^{(k)}[\mathbf{x}_i^{(k)}]\}_{k=1}^{M_i}= g_i\{\mathbf{h}_i^{(k)}\}_{k=1}^{M_i}
\end{equation}
The multimodal FL model of client $C_i$ can be expressed as:
\begin{equation}
\mathcal{F}_i(\mathbf{w}_i;\mathbf{X}_i) = \left\{\mathbf{w}_i; g_i\{\mathbf{h}_i^{(k)}\}_{k=1}^{M_i}\right\}
\end{equation}
where $\mathbf{w}_i$ denotes the learnable weights of the model associated with client $C_i$. Thus, the objective of multimodal FL is to train a set of models $\{\mathcal{F}_i(\cdot)\}_{i=1}^{N}$.
However, the heterogeneous model architectures result in learnable weights \( \mathbf{w}_i \) for each local model having different formats. Thus, these heterogeneous local model weights \( \{\mathbf{w}_i\}_{i=1}^{N} \) cannot be directly aggregated using existing FL methods. 

More explicitly, our objective in Eq.~\eqref{eq:eq5} is to minimize the loss function for clients with modality heterogeneity. This problem diverges from the conventional FL problem formulation in Eq.~\eqref{eq:eq1}.
\begin{equation}
\underset{\mathbf{w}_1,\mathbf{w}_2,...,\mathbf{w}_N}{\arg\min} \frac{1}{N}\sum_{i=1}^{N}\mathcal{L}\{\mathbf{w}_i; g_i\{{f_i}^{(k)}[\mathbf{x}_i^{(k)}]\}_{k=1}^{M_i}, y_i\}
\label{eq:eq5}
\end{equation}
\subsection{FedMM}
To tackle the issue of clients with heterogeneous but possibly overlapped modalities data, FedMM aims to train multiple single-modal feature extractors federatedly to help improve subsequent classification performance.
This contrasts with other methods focusing on training a multimodal fusion model. 
In FedMM, all clients collaboratively contribute to training multiple federated single-modal feature extractors. Subsequently, each client can utilize these federated extractors to extract features and perform classifications locally.

Specifically, for a multimodal client \(C_i\) with $M_i$ modalities, it parallelly trains a total of \(M_i\) distinct single-modal feature extractors \(\{f_i^{(k)}(\cdot)\}_{k=1}^{M_i}\). 
Beyond using the local true label for training, we innovatively incorporate the global prototype as a \textit{``pseudo-label''}.
This innovation allows us to design a dynamic loss function and overcome the challenge of updating federated feature extractors in the absence of labels due to privacy constraints.

The global prototype acts as a proxy for a given class and is determined by averaging the embeddings within the same class and modality across all clients \cite{tan2022fedproto}, \cite{Huang_2023_CVPR}, \cite{Chikontwe_2022_CVPR}. 
Notably, each modality has the same number of prototypes as the number of classes, and we omit to label the number of prototypes of the same modality below for brevity. The prototype of $k$-th modality can be expressed as:
\begin{equation}
\bar{\mathbf{p}}^{(k)} = \frac{1}{N_k} \sum_{i=1}^{N_k} \left\{\mathbf{h}_i^{(k)} \right\}
\end{equation}
Here, \( N_k \) is the number of clients that possess the \( k \)-th modality (i.e., a subset of \(N\) clients). Notably, federated feature extractors can be trained using both local true label \( \hat{y}_i \) and global prototype \( \bar{\mathbf{p}}^{(k)} \). 

The aim of training the feature extractor for the \( k \)-th modality is to minimize a loss function consisting of two components. The first term \( \scalebox{0.8}{\(\circled{1}\)} \) calculates the \( L2 \) distance loss between the local embedding \( \mathbf{h}_i^{(k)} \) and the global prototype \( \bar{\mathbf{p}}^{(k)} \). The second term \( \scalebox{0.8}{\(\circled{2}\)} \) incorporates the binary cross-entropy ($BCE$) loss between the local prediction \( \hat{y}_i \) and the local true label \( y_i \). The objective function is formulated as:
\begin{equation}
\small
\begin{aligned}
  \min \left\{ \underbrace{\lambda(t) \frac{\beta}{D} \left[ L2\left( \mathbf{h}_i^{(k)}, \bar{\mathbf{p}}^{(k)} \right) \right]}_{\clap{\scalebox{0.8}{\(\circled{1}\)}}} + \underbrace{\Bigl[1-\lambda(t)\Bigr] \Bigl[ BCE\bigl( \hat{y}_i, y_i \bigr) \Bigr]}_{\clap{\scalebox{0.8}{\(\circled{2}\)}}} \right\}
\end{aligned}
\label{eq:eq7}
\end{equation}
In more explicit terms, Eq.~\eqref{eq:eq7} is given by:
\begin{align}
  &\min \left\{ \lambda(t) \frac{\beta}{D} \left\| \mathbf{h}_i^{(k)} - \bar{\mathbf{p}}^{(k)} \right\|_2 \right. \nonumber \\
  &\left. + \Bigl[1-\lambda(t)\Bigr] \Bigl[ -y_i \log(\hat{y}_i) - (1-y_i) \log(1-\hat{y}_i) \Bigr] \right\}
\end{align}
Here, \( D \) represents the dimensionality of the embedding, while \( \beta \) is introduced to balance the magnitudes of the \( L2 \) and $BCE$ losses. Initially, it can be set to 1 and fine-tuned based on the losses observed during initial training. The dynamic weight \( \lambda(t) \) is defined as:
\begin{equation}
\lambda(t) = \frac{1}{1+e^{-\alpha(t-t_0)}}
\end{equation}
where \( t_0 \) designates the transition point (round) for shifting emphasis from the $BCE$ loss to the \( L2 \) loss, and \( \alpha \) modulates the sharpness of this transition.
\begin{figure}
\centering
  \includegraphics[width=0.52\textwidth, trim=230 90 180 60,clip]{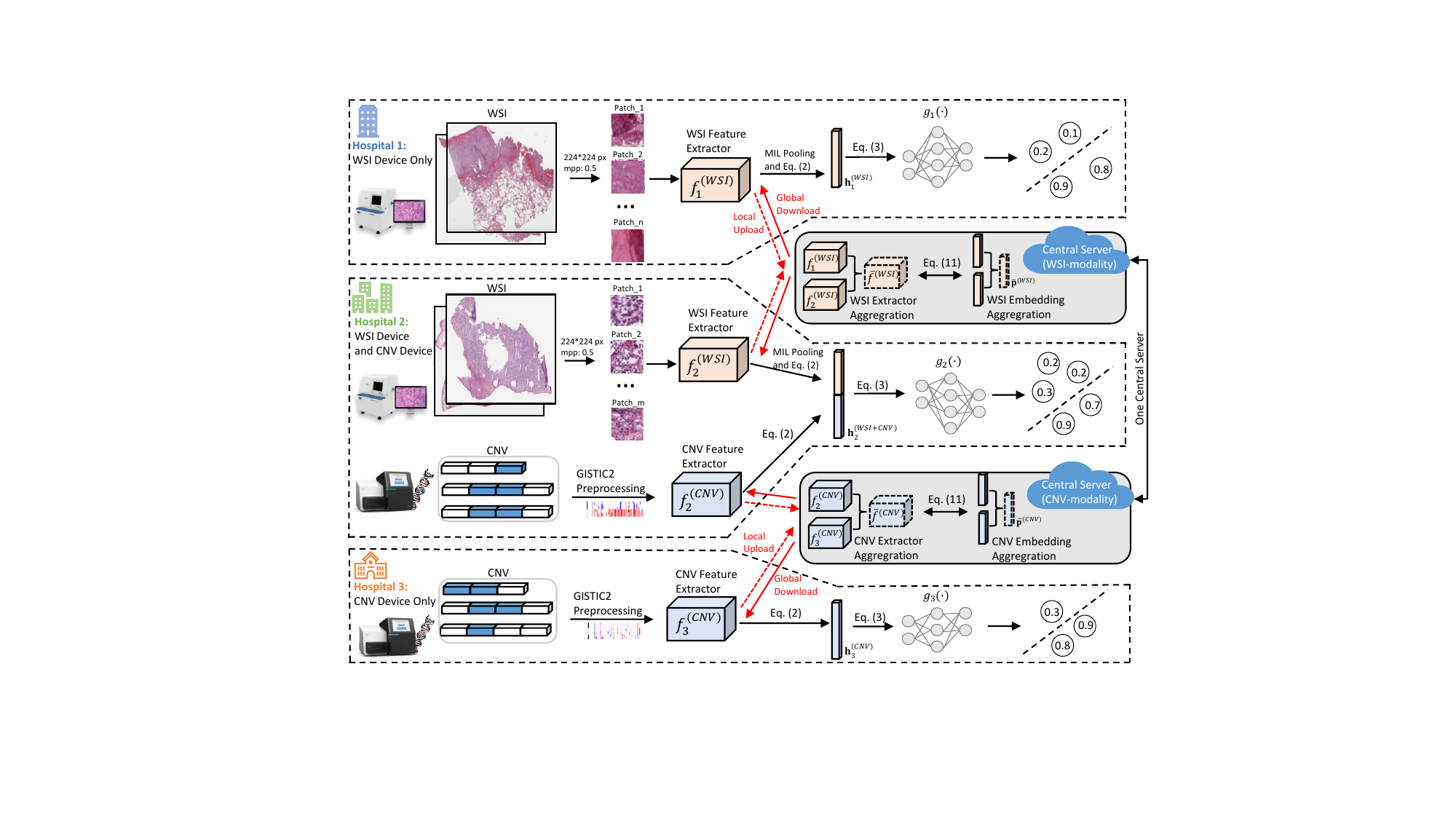}
  \caption{An illustration of a multimodal FL system with modality heterogeneity. It includes $N=3$ clients and one central server with $M=2$ modality processing components.
  }
\label{fig:2_FedMM}
\end{figure}
During training, it's crucial to dynamically adjust these losses using $\lambda(t)$, especially as the initial prototype might not accurately capture the actual underlying patterns of the data.
Giving too much weight to the \( L2 \) loss early on could make the model overly sensitive to these potentially inaccurate prototypes. 

Hence, $\lambda(t)$ starts by giving more emphasis to the $BCE$ loss, leveraging the local true label for clearer feedback.
As training progresses and the feature extractor refines its representations, the prototypes likewise become more accurate and reliable. At this point, $\lambda(t)$ gradually increases the emphasis on the \( L2 \) loss. 
More specifically, the first term \( \scalebox{0.8}{\(\circled{1}\)} \) aims to ensure the trained feature extractor generates embeddings closely aligned with the global prototype, and the second term \( \scalebox{0.8}{\(\circled{2}\)} \) aims to ensure the trained feature extractor can help make prediction values closer to the label on the local client.

More explicitly, in the \( t \)-th round of FedMM training, the procedures for client update and server update are as follows:

\textbf{Client Update:} Client \( C_i \) parallelly updates $M_i$ single-modal feature extractors $\{f_i^{(k)}\}_{k=1}^{M_i}$ using its local multimodal data \( \{\mathbf{x}_i^{(k)}\}_{k=1}^{M_i}\) and the received global prototypes \( \{\bar{\mathbf{p}}^{(k)}\}_{k=1}^{M_i} \) from the server.
\begin{equation}
f_i^{(k),t}(\cdot) \gets SGD\left\{ f_i^{(k),t-1}(\cdot), (\mathbf{x}_i^{(k)}, y_i, \bar{\mathbf{p}}^{(k)}) \right\}
\end{equation}

\textbf{Server Update:} The server runs \( 2 \times M \) threads performing model aggregation to update single-modal feature extractors and embedding aggregation to update global prototypes. Specifically, upon receiving the model weights from \( N_k \) clients, the server performs the extractor aggregation and embedding aggregation as follows:
\begin{equation}
\begin{cases}
\begin{aligned}
\bar{f}^{(k),t}(\cdot) &\gets \text{$Fed$}\left\{ f_1^{(k),t}(\cdot), f_2^{(k),t}(\cdot), \ldots, f_{N_k}^{(k),t}(\cdot) \right\} \\
\bar{\mathbf{p}}^{(k),t} &\gets \frac{1}{N_k} \sum_{i=1}^{N_k} \left\{ \mathbf{h}_i^{(k),t} \right\}
\end{aligned}
\end{cases}
\label{eq:eq10}
\end{equation}
Here, \(Fed(\cdot)\) can encompass any established FL methods, including but not limited to FedAvg \cite{mcmahan2017communication} and FedProx \cite{li2020federated}. This flexibility allows FedMM to seamlessly integrate with various state-of-the-art FL methods, thereby enhancing its potential for real-world applications.

Subsequently, these federated feature extractors are deployed locally to extract corresponding modality features from small-scale, privacy-sensitive datasets on each client. These extracted features then serve as the input for the local classifiers residing on each client.

In sum, this transformation from a complex and heterogeneous multimodal input into multiple manageable single-modal inputs provides an effective solution to the challenges posed by heterogeneous modalities within a multimodal FL system. 

\vspace{-2mm}
\section{Experiments}
\label{sec:exp}
\subsection{Datasets}
We evaluated 
FedMM using two publicly available datasets from The Cancer Genome Atlas (TCGA)
\footnote{https://portal.gdc.cancer.gov/repository}. 
To mirror the real-world challenge of modality heterogeneity across various hospitals, we reasonably distributed these data among three hospitals,
as shown in Table ~\ref{tab:table1}.
The flow of the experiment is illustrated in Fig.~\ref{fig:2_FedMM}.

\textbf{TCGA-NSCLC}: This dataset contains data from patients with non-small cell lung cancer (NSCLC).
It contains two subtypes: adenocarcinoma (LUAD) and squamous cell carcinoma (LUSC).

\textbf{TCGA-RCC}: This dataset includes data from patients with renal cell carcinoma (RCC), the most prevalent type of kidney cancer. 
It mainly includes three subtypes: Kidney Renal Clear Cell Carcinoma (KIRC), Kidney Renal Papillary Cell Carcinoma (KIRP), and Kidney Chromophobe (KICH). 
Due to limited KICH samples, we focus solely on binary classification for KIRC and KIRP subtypes.
\begin{table}
\centering
\caption{Sample number details of two public datasets from TCGA.} 
\resizebox{\columnwidth}{!}{
\begin{tabular}{c|cccccc} 
\hline
Dataset & Modality & Class & Hospital 1 & Hospital 2 & Hospital 3 & Total \\
\hline
\multirow{4}{*}{TCGA-NSCLC} & \multirow{2}{*}{WSI} & LUAD & 199 & 315 & 0 & \multirow{2}{*}{1009} \\
                            &                      & LUSC & 210 & 285 & 0 & \\
\cline{2-7} 
                            & \multirow{2}{*}{CNV} & LUAD & 0 & 315 & 203 & \multirow{2}{*}{1022} \\
                            &                      & LUSC & 0 & 285 & 219 & \\
\hline 
\multirow{4}{*}{TCGA-RCC}   & \multirow{2}{*}{WSI} & KIRC & 246 & 291 & 0 & \multirow{2}{*}{824} \\
                            &                      & KIRP & 129 & 158 & 0 & \\
\cline{2-7} 
                            & \multirow{2}{*}{CNV} & KIRC & 0 & 291 & 243 & \multirow{2}{*}{825} \\
                            &                      & KIRP & 0 & 158 & 133 & \\
\hline
\end{tabular}
}
\label{tab:table1}
\end{table}

\subsection{Implementation and Reproducibility}
\textbf{WSI Feature Extractor}: Owing to the gigapixel size of the WSI (e.g., \(40,000\times40,000\) pixels), using a CNN directly for feature extraction was impractical. To overcome this, we adopted a Multiple Instance Learning (MIL) strategy \cite{carbonneau2018multiple}. We cropped the WSI into multiple non-overlapped patches, each measuring \(224 \times 224\) pixels.
After removing the background patches, we used ResNet-34~\cite{He_2016_CVPR} for patch feature extraction. Specifically, we froze the preceding layers, which were pre-trained on ImageNet~\cite{deng2009imagenet}, while fine-tuning the last convolutional layer in the third residual block.
Importantly, tumor regions were notably large, making up over 80\% of the full WSI in both the TCGA-NSCLC and TCGA-RCC datasets. 
This allowed us to effectively aggregate patch-level representations, generating WSI-level representation \( \mathbf{h}_i^{\text{(WSI)}} \) using appropriate pooling methods like attention pooling \cite{ilse2018attention}, \cite{wang2019comparison}, which was commonly used in MIL.

\textbf{CNV Feature Extractor}: Initial preprocessing of the CNV data was performed using GISTIC2 \cite{mermel2011gistic2}. We then used the Self-Normalizing Network (SNN) \cite{klambauer2017self} comprised of two hidden layers to tackle the issues of high-dimensional genomic data with limited samples \cite{chen2020pathomic}. The final fully connected layer was responsible for learning the representation \( \mathbf{h}_i^{\text{(CNV)}} \) \cite{chen2022pan}.
The experiments involved three clients and one central server. The training was executed on one Nvidia Tesla V100 GPU and three Nvidia Tesla P100 GPUs. We set global training rounds to 100 and the local epoch to 3, with all clients participating in each round. 
The learning rate and batch size were set to 0.001 and 32. $\beta$, $\alpha$, and $t_0$ were set to 0.25, 0.05 and 30. We repeated the experiment 20 times, randomly splitting the dataset into training and testing at 0.8:0.2.

\subsection{Experimental Results and Analysis} 
We evaluated performance using both accuracy and ROC curves for binary classification tasks, comparing FedMM against two baselines.

\textbf{Local Training (Baseline 1):} Each local model is trained independently, without considering modality overlaps among clients. 

\textbf{Multi-FedAvg (Baseline 2):} All modalities from all clients are considered. Missing modalities on local clients are set to zero.

\begin{figure}
\centering
\begin{subfigure}{0.475\linewidth}
  \includegraphics[width=\textwidth, trim=10 1 10 11,clip]{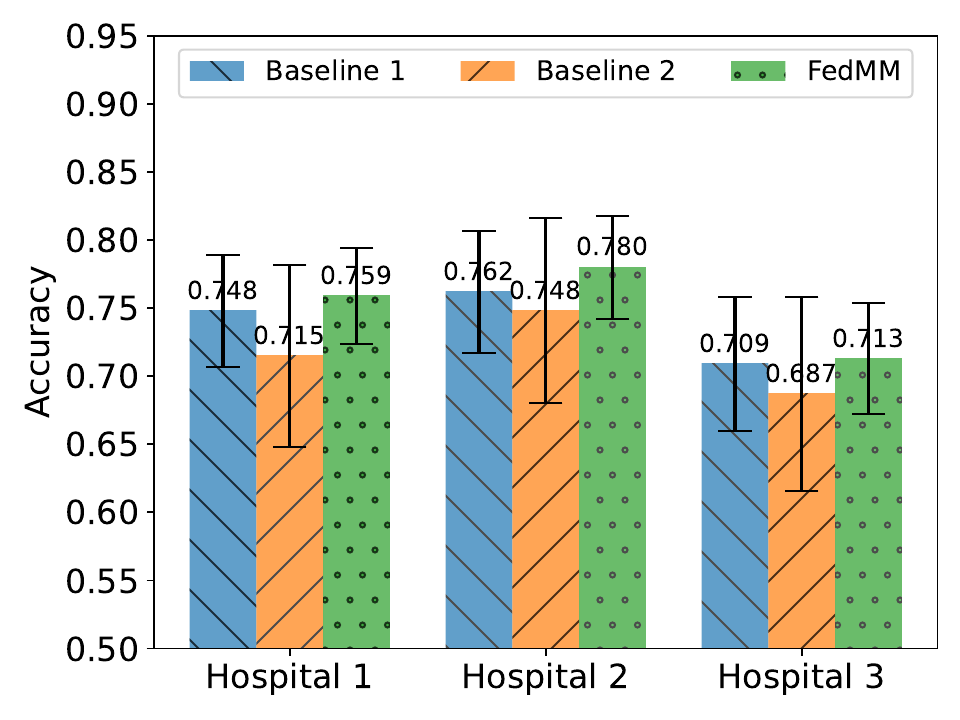}
  \caption{Accuracy (TCGA-NSCLC)}
\end{subfigure}
\hfill
\begin{subfigure}{0.515\linewidth}
  \includegraphics[width=\textwidth, trim=5 0 10 20,clip]{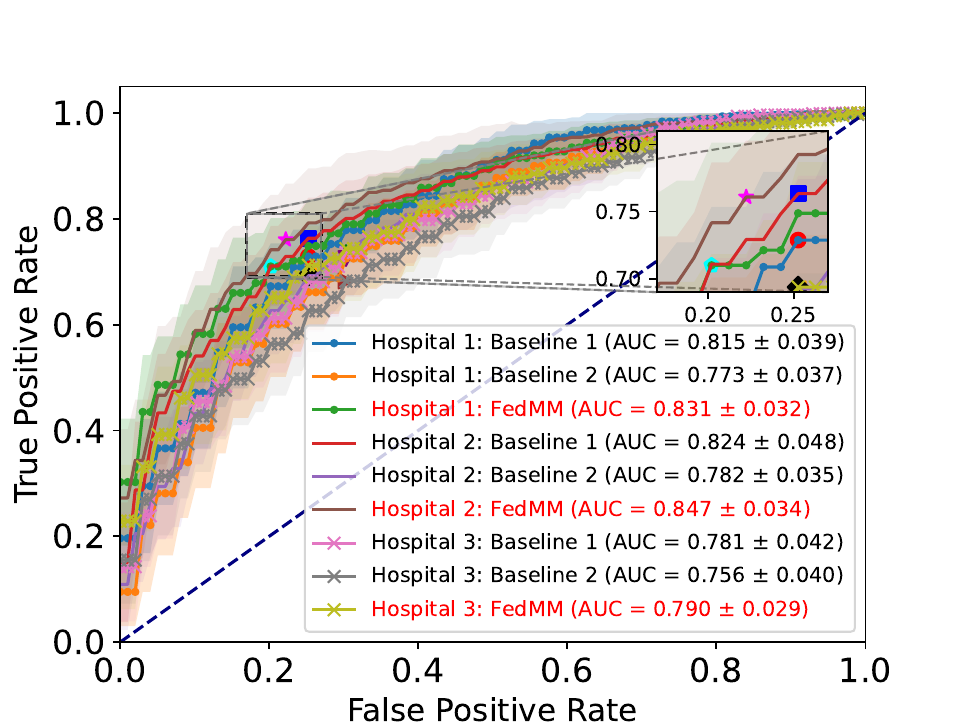}
  \caption{ROC Curve (TCGA-NSCLC)}
\end{subfigure}
\begin{subfigure}{0.475\linewidth}
  \includegraphics[width=\textwidth, trim=10 1 10 11,clip]{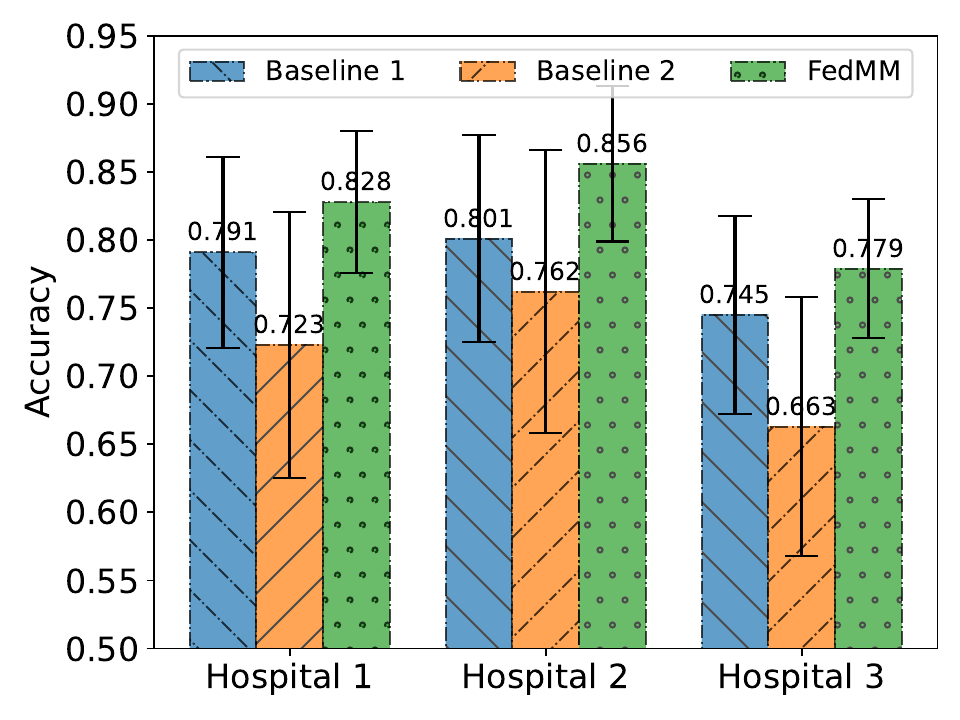}
  \caption{Accuracy (TCGA-RCC)}
\end{subfigure}
\hfill
\begin{subfigure}{0.515\linewidth}
  \includegraphics[width=\textwidth, trim=5 0 10 20,clip]{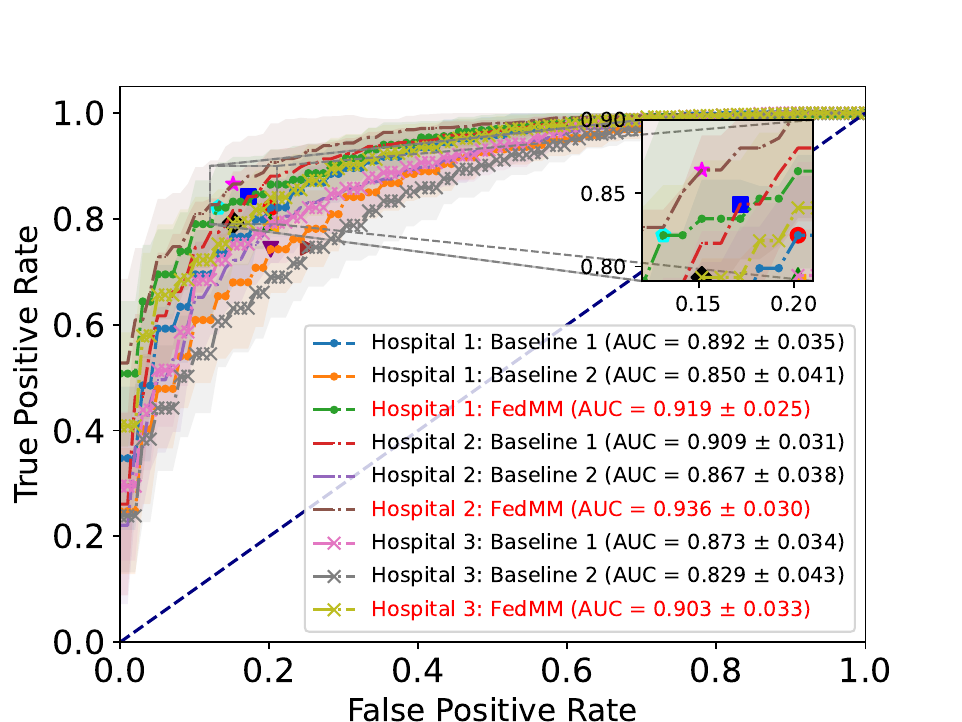}
  \caption{ROC Curve (TCGA-RCC)}
\end{subfigure}
\vspace{-3mm}
\caption{Comparison of the classification performance between FedMM and baselines based on two public datasets.}
\label{fig:4_results}
\end{figure}

FedMM significantly outperforms two baselines across three hospitals, as shown in Fig.~\ref{fig:4_results}. For example, for the TCGA-NSCLC dataset at hospital 2, FedMM surpasses local training and Multi-FedAvg baselines by AUC values of 0.023 (2.79\%$\uparrow$) and 0.065 (8.31\%$\uparrow$), respectively. Similarly, for the TCGA-RCC dataset at the same hospital, FedMM exceeds two baselines by AUC values of 0.027 (2.97\%$\uparrow$) and 0.069 (7.96\%$\uparrow$). 
Notably, a direct comparison of FedMM and Multi-FedAvg highlights FedMM's effectiveness in tackling the issue of modality heterogeneity. 
Finally, while FedMM outperforms all baselines with similar training sample sizes (e.g., in hundreds), collecting more data remains essential for distributing the data across more clients and further validating FedMM.
\vspace{-3mm}
\begin{figure}[H]
\centering
\begin{subfigure}{0.32\columnwidth}
  \includegraphics[width=\textwidth, trim=10 10 10 10,clip]{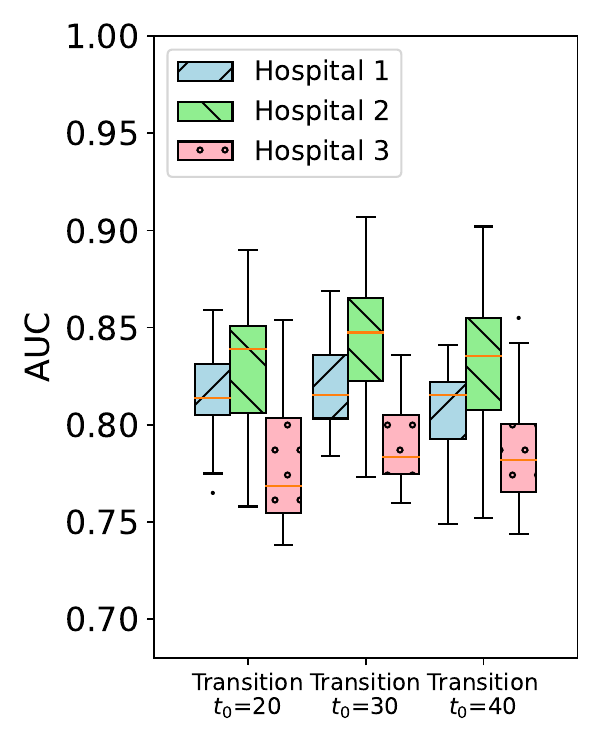}
  \caption{Loss Transition Point}
\end{subfigure}
\begin{subfigure}{0.32\columnwidth}
  \includegraphics[width=\textwidth, trim=10 10 10 10,clip]{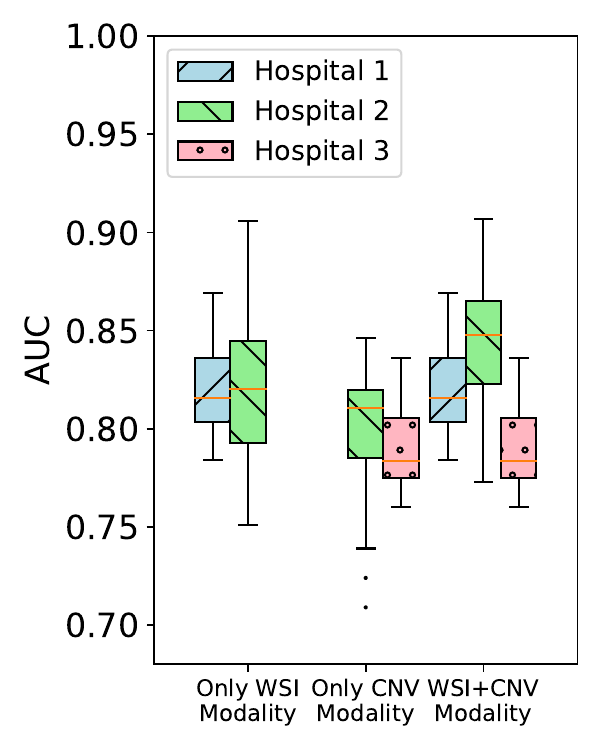}
  \caption{Modality Types}
\end{subfigure}
\begin{subfigure}{0.32\columnwidth}
  \includegraphics[width=\textwidth, trim=10 10 10 10,clip]{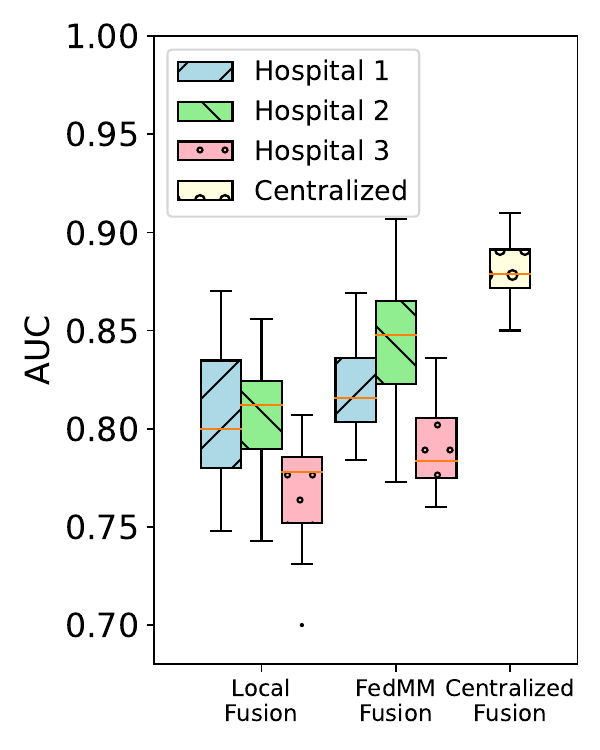}
  \caption{Fusion Mechanisms}
\end{subfigure}
\caption{Ablation study of FedMM on the TCGA-NSCLC dataset.}
\label{fig:5_AB}
\end{figure}
\textbf{Ablation Study:} 
To rigorously evaluate the impact of various components on the performance of FedMM, we further conduct a comprehensive ablation study, which investigates the transition point of the two losses, different modality inputs, and various fusion mechanisms.
We present results on the TCGA-NSCLC dataset, as shown in Fig.~\ref{fig:5_AB}, with similar results on the TCGA-RCC dataset.
Notably, centralized fusion is impractical in real-world situations due to privacy constraints on aggregating data from all hospitals.
However, the AUC value achieved by FedMM closely approaches that of centralized fusion, highlighting its practical effectiveness.

\vspace{-2mm}
\section{Conclusion}
\label{sec:conclusion}
In conclusion, we propose a FedMM framework to train multiple single-modal feature extractors federatedly to help improve subsequent classification performance. 
Hospitals can leverage these federated trained extractors to perform local feature extraction and classification tasks. 
Our comprehensive evaluations show that FedMM notably outperforms two baseline methods, proving its effectiveness in multimodal and privacy-sensitive computational pathology.


\vfill\pagebreak


\section{References}
\printbibliography

\end{document}